\documentclass{ecai}
\usepackage{graphicx}
\usepackage{latexsym}

\newtheorem{definition}{Definition}[section]
\usepackage{amsfonts, amsmath}
\usepackage{relsize}
\usepackage{algorithm, algpseudocode}
\usepackage{tabularx, multirow}

\begin{document}

\begin{frontmatter}

\title{Combination of Weak Learners eXplanations to Improve Random Forest
eXplicability Robustness}

\author[A]{\fnms{Riccardo}~\snm{Pala}}
\author[A]{\fnms{Esteban}~\snm{García-Cuesta}\orcid{0000-0002-1215-3333}\thanks{Corresponding Author. Email: esteban.garcia@upm.es}}

\address[A]{Departamento de Inteligencia Artificial, Universidad Politécnica de Madrid, Madrid, Spain}
%\address[A]{Departamento de Inteligencia Artificial, Universidad Politécnica de Madrid, E.T.S.I.I Campus de Montegancedo s/n, Boadilla del Monte, 28660, Madrid, Spain}

\begin{abstract}
The notion of robustness in XAI refers to the observed variations in the explanation of the prediction of a learned model with respect to changes in the input leading to that prediction. Intuitively, if the input being explained is modified slightly subtly enough so as to not change the prediction of the model too much —then we would expect that the explanation provided for that new input does not change much either. We argue that a combination through discriminative averaging of ensembles weak learners explanations can improve the robustness of explanations in ensemble methods. This approach has been implemented and tested with post-hoc SHAP method and Random Forest ensemble with successful results. The improvements obtained have been measured quantitatively and some insights into the  explicability robustness in ensemble methods are presented. \\

\textbf{keywords}: explainable AI, robustness of explanations, weak explanations, ensembles, model multiplicity
\end{abstract}

\end{frontmatter}

\section{Introduction}
There are many areas of competence where, for a machine learning (ML) model, the mere fact of producing accurate predictions is not enough to convince the end user to trust the application. In such cases, the problem of understanding how a model produces a given result may become a matter of great importance \cite{Goodman}. Many algorithms construct models that are opaque to humans \cite{Burkart}, but in fields such as medicine it is necessary to provide explanations that support the output of a model in order to be adopted~\cite{Tjoa}.
When a model does not meet the requirements to be considered intrinsically explainable, it is necessary to apply methods to explain the reasons for its decisions. In this case, we are referring to post-hoc eXplainable AI (XAI) techniques which, by following the taxonomy specified in \cite{Barredo}, can be divided into \emph{model-agnostic}, which are applicable to any type of model, or \emph{model-specific}, which are tailored or specifically created to explain certain ML models.
One of the most widely used algorithms is \emph{SHAP} (SHapley Additive exPlanations) \cite{Lundberg}, which is an XAI method to explain individual predictions and it is based on Shapley values originally used in game theory. It provides a way to assign an importance value to each feature based on its contribution to the prediction. In a ML setting, the feature values of a data instance act as players in a coalition. In such a case, the Shapley value is the average marginal contribution of a feature value across all possible partitions of the feature space, that can be a positive or negative value basing on the type of influence of the feature on the prediction. The final SHAP explanation for a given input $x$ is the set of all feature-specific SHAP explanations.

Although this XAI technique enjoys many desirable properties, including the lack of need for modifications to existing models in order to apply it (it is model-agnostic), it also suffers from several limitations, mainly concerning its robustness. Intuitively, robustness states that similar inputs must produce similar explanations, i.e, regardless of its method of representation the explanation must remain constant in its vicinity in order to be considered valid~\cite{Alvarez}. This lack of robustness has been also associated by~\cite{D'Amour} with the problem of \emph{underspecification} and it is identified as the cause of any poor capacity to provide credible explanations when applied to real-world domains. Quoting the paper, "\emph{the solution to a problem is underspecified if there are many distinct solutions that solve the problem equivalently}". What makes this parallelism fit is that the production of two radically distinct explanations derived from two very similar data points indicates that the algorithm believes that the (almost) same problem can be solved equivalently in two different ways. This overtakes the misconception of two premises: that accuracy is the only measure of how "good" a model is (usually, the model selection problem boils down to accuracy maximization), and that, for a given task, models that maximize accuracy don't differ significantly from each other. The authors in~\cite{Black} refers to this behavior as \emph{model multiplicity} where several models for a given prediction task have equivalent accuracy, yet different in their model internals. In that work the authors also point out that, although Bayes optimal predictors are unique, multiplicity occurs because the learned models are far enough away from the Bayes optimum to leave plenty of room for multiplicity. This has been also proved experimentally at other works such as~\cite{Mehrer}. 

On the other hand, ensembles of models, i.e. models resulting from the aggregation of the contribution of several so-called \emph{weak learners} have some desirable properties to provide more accurate and robust predictions. In the case of classifiers, ensembles are designed to increase the generalization capabilities of the single models by training several of them and combining their decisions to obtain a single class label~\cite{Galar}. Particular attention is placed on \emph{Random Forest}~\cite{Breiman2001, Ho} that is a very successful machine learning tool that exploits the combination of independent decision trees to build up a more powerful learner that is able to mitigate some problems such as outliers. Random Forest is able to provide a measure of the importance of each feature regarding the prediction, however this is not sufficient to consider the model transparent and this intrinsic explainability decreases as the number of weak learners in the ensemble increases.  It is, in fact, necessary to have a better understanding of how much each feature contributes to a certain label assignment and especially to understand it on a sample by sample basis. One way to overcome the inherent lack of explainability of Random Forest models is to apply XAI model-agnostic algorithms such as SHAP. At~\cite{Alvarez} the authors observed that such an application yields values that only partially meet the expectations. Although the method provides the contribution of every feature for each data point in most cases the corresponding values have low robustness to small perturbations of the input. This can be interpreted as a symptom of a lack of trustworthiness in these explanations. In addition, Decision Tree is a model with very good intrinsic explainability, but by design it creates hard decision boundaries meaning that small changes in the input can lead to abrupt changes in the explanations. Random Forest relies on the combination of several weak learners to create a smoother decision boundary that better adapts to the real one. Hence, it is expected that the softer boundaries that provide a more robust model will also provide more robust explicabilities.

This work is devoted to the development of an efficient and effective method for the application of model-agnostic XAI techniques to ensemble models. The objective is to  exploit the prediction capabilities of ensembles and improve their robustness to enable the production of  more trustable explanations. This means that, given a certain data point $x$ and a slightly perturbed version of it $x'$, we expect the explanations $y$ and $y'$, respectively produced from the two inputs just mentioned, to differ marginally. For this purpose, we take advantage of the decomposability of the Random Forest ensemble model in order to exploit the explanation values provided by each single weak learner to the purpose of building a more robust global explanation by combination of those weak explanations. \\

\textbf{RF-specific Explainaibility Methods.} We include here a specific part to discuss the inherent explainability of Random Forest in order to provide a global overview of efforts done on XAI specifically for this technique. The following are some examples of the current state of the art explainability techniques in the domain of Random Forest and, more generally, tree ensembles, mostly taken from \cite{Ribeiro}. Starting from methods of \emph{simplification}, in \cite{Fernandez} counterfactual sets are extracted from the model to create a more transparent version of the same, \cite{Domingos} poses the idea of training a less complex model, and thus more intrinsically explicable, on samples randomly extracted from the test set labelled by the ensemble, in \cite{Deng} it is explained how to construct a Simplified Tree Ensemble Learner (STEL) on the basis of rules extracted from the ensemble and selected through feature selection and complexity-driven criteria, finally \cite{Hara} presents as a solution that of training two models, one more complex (i.e. opaque) dedicated to prediction and a simpler one (i.e. transparent) that will be used to extract explanations, whose simultaneous use is governed by statistical divergence measures. Speaking, instead, of methods related to feature importance, among the earliest works on the subject we can find \cite{Breiman1984} and \cite{Archer} in which the influence of a feature is analysed by means of random permutations of Out-Of-Bag (OOB) samples and measured through the use of metrics such as MDA (Mean Decrease Accuracy) and MIE (Mean Increase Error), which is followed by the work in \cite{Auret} that makes use of feature importance measurement and partial dependency plots as tools to provide humans with information regarding the underlying learning processes in order to successfully extract knowledge from them.
Nevertheless, although these can be considered successful applications in the field of tree ensembles' explainability, in all cases it can be seen that robustness is not a property that is contemplated as a measuring instrument for their effectiveness.

The following section presents the proposed method and conduct of the experiments and the data used. Then, the experimental results and the main inferred advantages and disadvantages are shown. Finally, some conclusions are presented.

\section{Robustness of eXplicability.}
By examining the literature, it can be observed that robustness is a property for which a measurement method has not yet been unanimously defined. Furthermore, among the various works that can be found on this subject, very few concern the specific application of the concept of robustness to model explanations. In this work we make use of the notion of \emph{local Lipschitz continuity} to conduct the experiments. Before going into details of robustness definition we need to introduce that we decided to conduct the experimental part of our proposed method using Decision Tree and Random Forest as the reference models, and SHAP as the reference XAI technique. The main reason for this choice is their simplicity as a key feature in understanding the mechanisms behind the process of evaluating the robustness of ensemble models explanations. It was possible in this way to relate the results back to some well-known behaviors of the models and method, so as to provide better insights into the proposed method.  \\

\begin{figure}[!h]
\centering
\includegraphics[width=0.9\columnwidth]{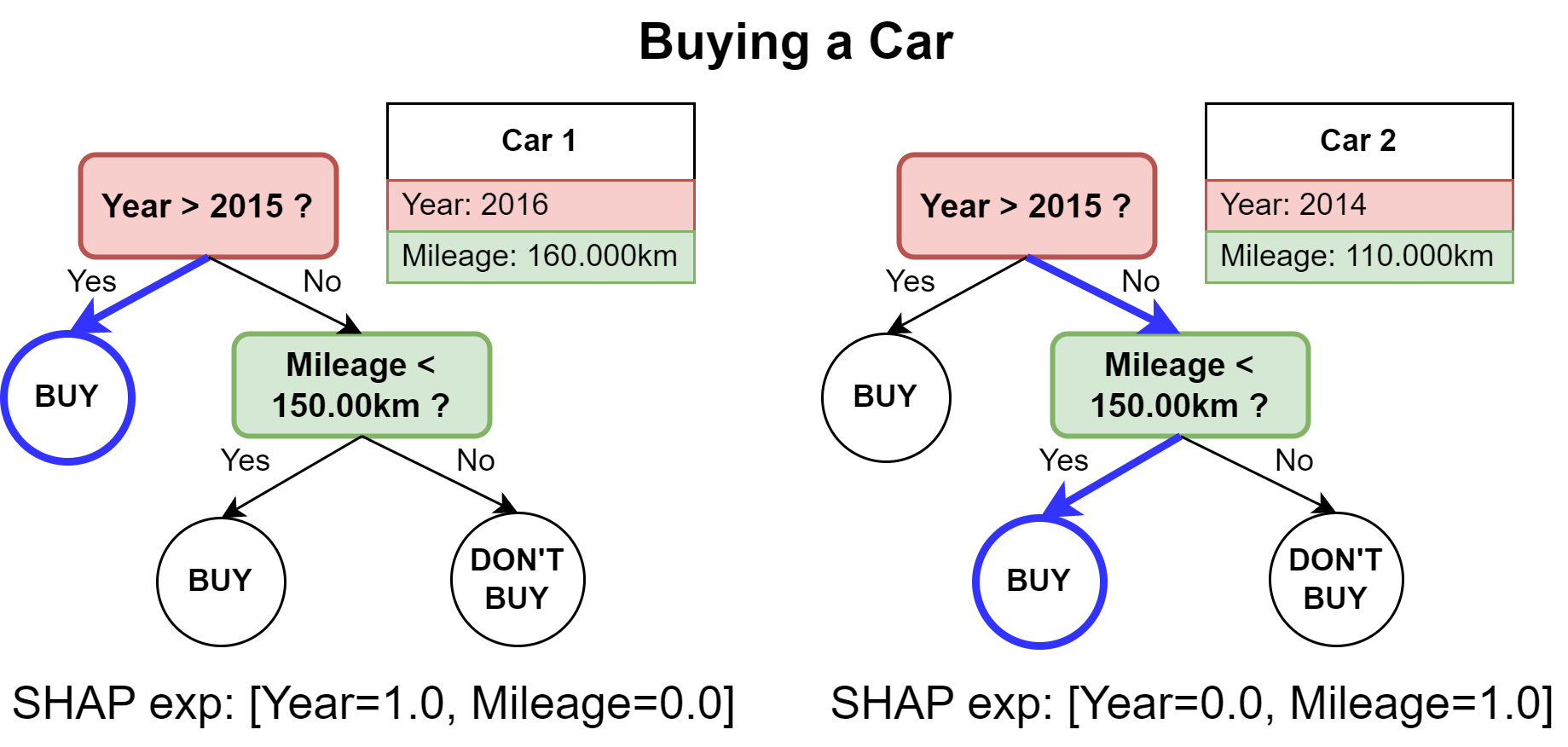}
\vspace{-5mm}
\caption[Regions of explanation constancy]%
{If two data points activate the same prediction branch they will produce the same SHAP explanation values, while two data points that activate different decision branches will lead to the production of different ones. The latter case doesn't necessarily imply that the predicted labels are different. This  illustrates how two inputs that share the same prediction label can lead to the production of two different explanations.}
\label{fig:figure11}
\end{figure}

\textbf{Regions of explanation constancy.}\label{zodc} SHAP is a method designed in such a way that, when applied to Decision Tree, as long as one feed the algorithm with input data that activate the same decision branch, it will always provide constant explanation values. We will refer to the portions of space within which the XAI method outputs a constant value as \emph{regions of explanation constancy}. When dealing with tree ensembles, two data points $x_i$ and $x_j$ do \textbf{not} fall in the same zone of explanation constancy if the two inputs activate two different decision branches in at least one of the weak trees of the ensemble. Note that this does not necessarily mean that the two input points produce a different prediction (Figure \ref{fig:figure11}. shows an example of this type of behavior).\\

\subsection{Robustness Metric Definition}\label{sec:metric}
To quantify the robustness property of an explanation we follow the presented in \cite{Alvarez}, fell on the notion of \emph{Lipschitz continuity}, and defined as follows:
\begin{definition}
    $f : \mathcal{X} \subseteq \mathbb{R}^n \rightarrow \mathbb{R}^m$ is \textbf{locally Lipschitz} if for every $x_0$ there exist $\delta>0$ and $M \in \mathbb{R}$ such that $||x - x_0||<\delta$ implies $||f(x) - f(x_0)||\leq M||x - x_0||$.
\end{definition}
Making use of this notion, the paper evaluates the robustness $\hat{L}$ as the maximum variation of the explanation value $g(x_i)$ for each sample $x_i$ of the test set, being $B_{\epsilon}(x_i)$ a circle of radius $\epsilon$ around the data point in search $i$:
\begin{equation}\label{eqLipschitzMax}
    \hat{L}(x_i) = \max_{x_j \in B_{\epsilon}(x_i)} \frac{||g(x_i) - g(x_j)||_2}{||x_i - x_j||_2}
\end{equation}
where the function $g(\cdot)$ of interest is the one implemented by the XAI algorithm, in our case the SHAP function.

\begin{figure}[!t]
\centering
\includegraphics[width=\columnwidth]{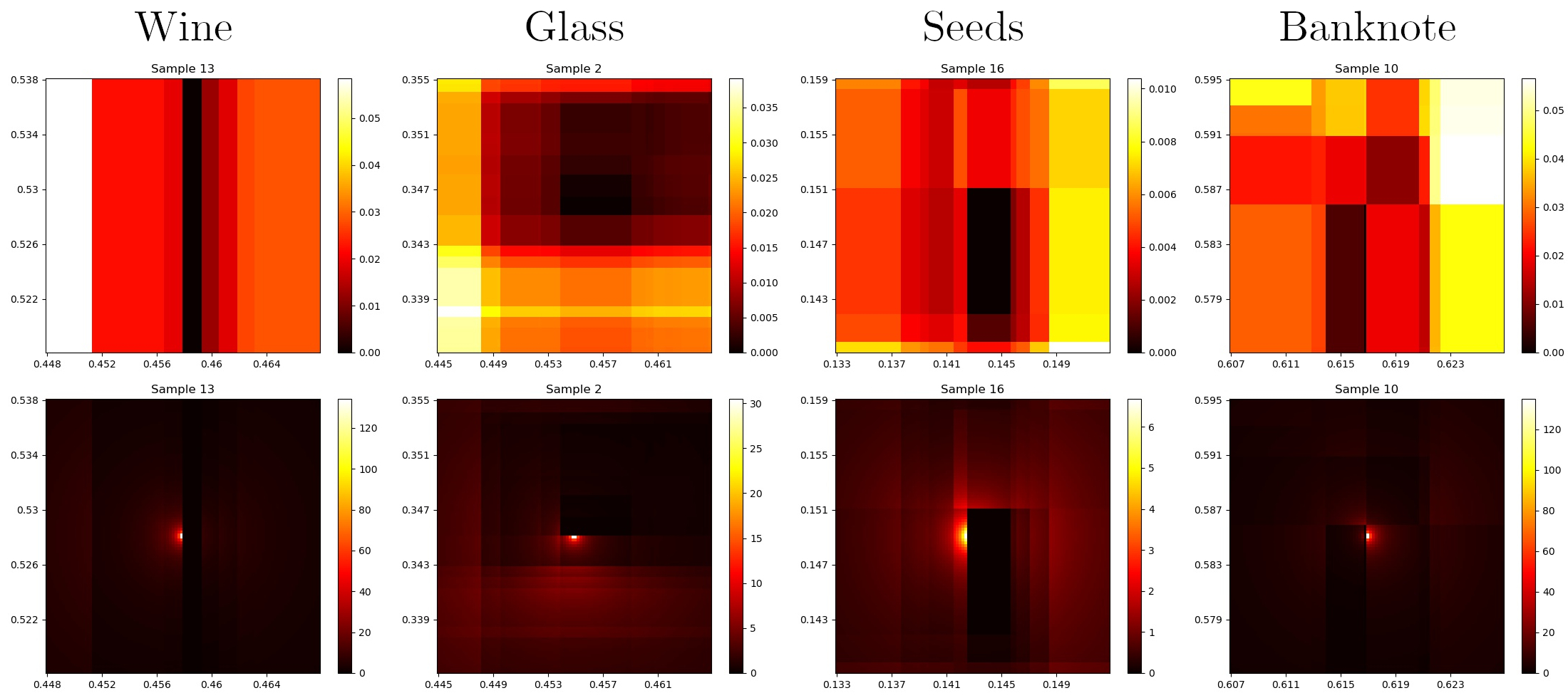}
\vspace{-5mm}
\caption[SHAP explanations difference and incremental ratio heatmaps comparison]%
{\textbf{SHAP explanations robustness criteria problem} - Top row shows the eq. (\ref{heatmapsDifference}) plots and bottom row the eq. (\ref{heatmapsRobustness}) of the SHAP explanations produced by Random Forest in a neighbourhood of some $x_i$ points of interest, on each of the four datasets. At the bottom row it is observed how proximity to the centre of the search space is a factor that leads to the production of high values whereas the differences between in the surrounding areas are obscured.}
\label{fig:figure1}
\end{figure}

The choice of searching for the maximum value results in measurements that are unreliable for the purpose of a balanced and fair calculation of robustness around a given data point. The reason lies in the fact that, especially in the specific case where SHAP explanations are applied to a Random Forest model (composed by several weak Decision Trees), the values of interest are eclipsed by the peak measured in the area closest to $x_i$. Within each~\emph{region of explanation constancy} (see discussion at the beginning of Section \ref{zodc}) different from the one to which the $x_i$ point belongs to, the value of fraction in (\ref{eqLipschitzMax}) increases as the distance of $x_j$ with $x_i$ decreases, because of the denominator. In other words, when $||x_i - x_j||_2$ has a very low value, but large enough to change decision branch even in only one of the weak trees in the ensemble, the value of $\hat{L}(x_i)$ diverges.This is a very frequently encountered case as the number of estimators increases. 
%also basing on the number of estimators of which the ensemble is composed, just consider the case when the data point $x_i$ around which we want to compute the robustness of our algorithm, is on the boundary of one of the above-mentioned zones. 
Fig. \ref{fig:figure1} shows, for each of the examined datasets, an example comparing the values obtained by first considering the difference of the SHAP explanation values (top) and then their incremental ratio (bottom). All the heatmaps are produced basing the values obtained from 10000 perturbed $x_j$ points, 100 for each row and 100 for each column and they represent  two possible dimensions chosen based on its significance and predictions stability to the perturbations (which produce areas of interest for the purposes of the robustness calculation). Hence, each $x_m$ point of the map represents a 2-dimensional version of the $p$-dimensional point $x_j$. Being $a_x$ and $a_y$ the two features that vary along the two axes, we define the correspondence between $x_j$ and $x_m$ as:
\begin{align*} %\label{xjmCorr}
    x_j(x_m) = \{ & x_{j,1},...,x_{j,a_x-1},x_{m,a_x},x_{j,a_x+1}, \\
    & ...,x_{j,a_y-1},x_{m,a_y},x_{j,a_y+1},...,x_{j,p} \}
\end{align*}
In this way we can formally define the computation of the \emph{heat} value of each point of the two maps. The explanations' difference map follows:
\begin{equation} \label{heatmapsDifference}
    H_d(x_m) = ||g(x_i) - g(x_j(x_m))||_2
\end{equation}
while the explanations' incremental ratio map follows:
\begin{equation} \label{heatmapsRobustness}
    H_r(x_m) = \frac{||g(x_i) - g(x_j(x_m))||_2}{||x_i - x_j(x_m)||_2}
\end{equation}
where, again, $g(\cdot)$ is the SHAP explanation function.
It can be observed that the tendency of the values in the incremental ratio is towards infinity as one approaches the centre of the space whereas the differences in SHAP explanation are sometimes negligible. This behavior is the consequence of the presence of the normalization factor $||x_i - x_j||_2$ at the denominator of (\ref{eqLipschitzMax}), which value tends to 0 as $x_j$ approaches $x_i$. This is reflected in the fact that using such a metric concentrates in penalizing the models that creates many decision/explanation boundaries rather than assessing actual robustness within the neighborhood.

Nevertheless, it still remains necessary to take into account the distance between points to balance the more distant $x_j$ points. Therefore, we propose a slightly modified version of eq.~(\ref{eqLipschitzMax}) to calculate the \textbf{average} value of the incremental ratio between the explanation of $x_i$ and the explanation of the $x_j$ points around it as the new SHAP robustness criteria:

\begin{equation}\label{eqLipschitzAvg}
    \bar{L}(x_i) = \frac{1}{|\mathcal{N}_{f,\epsilon}(x_i)|}\sum_{x_j \in \mathcal{N}_{f,\epsilon}(x_i)} \frac{||g(x_i) - g(x_j)||_2}{||x_i - x_j||_2}
\end{equation}

being $\mathcal{X}$ the input space to which all $x_i$ points belong, $\mathcal{A}$ the set of features of $\mathcal{X}$, $f(\cdot)$ be the prediction function of the model, and for every $x_i$ sample of the considered test set, a discriminative discretization of its surrounding, in which points are evenly distributed, as:
\begin{equation*} %\label{eqNeighborhood}
    \mathcal{N}_{f,\epsilon}(x_i) = \left\{ x_j \in \mathcal{X}~ \mathlarger{|}~ \left |x_{i,a} - x_{j,a} \right | \leq \epsilon~\forall a \in \mathcal{A},~ f(x_i)=f(x_j) \right\}
\end{equation*}
being $x_{i,a}$ the value of the feature $a$ of data point $x_i$.

It can be seen that the computation of this value is significantly lighter than the one defined in eq. (\ref{eqLipschitzMax}) while preserving a good exhaustiveness of the analysis of the surrounding of the point considered.
It is important to note that this function provides robustness values in arbitrary units and thereof it can't be interpreted directly as a quality value. They have to be interpreted in comparison with other results. For instance, in a comparative analysis between the robustness of explanations of different models applied to the same dataset using the same XAI technique.

%On top of that, as one can notice from the definition of the neighborhood defined above, we decided to include in the robustness calculation only the perturbed samples whose label predicted by the model is the same as the original sample. The reason behind this choice is that, intuitively, we only expect robust explanation values as long as these only account for a single output value. Indeed, it is reasonable to expect that a perturbed data point whose label differs from that of the original data point will produce an explanation that differs substantially from that of the original data point, since different outputs are understood with different explanations. To fix the ideas, given $f(\cdot)$ and $g(\cdot)$, respectively the model's prediction function and explanation function, taken a point $x_i$ and a perturbed version of it $x_j$, if $f(x_i) \neq f(x_j)$ then we expect $g(x_i)$ and $g(x_j)$ to also differ substantially. Thus, considering perturbed samples with a label different from the original one in the robustness calculation would lead to an unfair robustness calculation, which would reward algorithms that produce robust values when this property is undesirable.

\subsection{Combination of Weak eXplanations}

As introduced earlier, the robustness of the explanations turns out to be a crucial property for ensembles to be considered reliable and trustworthy. Intuitively, if an ensemble of models performs well on a given dataset, it is then likely that most of the weak learners of which it is composed will contain extractable knowledge that can be considered useful in terms of explaining a certain decision. Indeed, producing explanations coming from the output of an ensemble of several weak learners is desirable for humans to understand a prediction, for instance, many people feel more confident about a medical opinion when it is the result of the union of the opinions of multiple experts.
Ensembles have been widely shown to increase the robustness of predictions and we expect that this can be transferred to explanations as well. SHAP~\emph{regions of explanation constancy} play a key role to make this transfer. These regions cause abrupt changes in the SHAP explanations provided by individual decision tree models (weak learners), because the low complexity of the decision branches structure brings to less frequent, although more sudden, change in explanation values. On the other hand, we expect that the decision boundaries associated to an ensemble will be smoother (due to the \emph{overlapping}) and thereof will led to the production of smoother SHAP explanations as well.  This can be proved at Random Forest by analyzing the produced explanations that are the result of such overlapping, we present here a notationally-adapted version of SHAP formula, for a multi-labelled classification setting:
\begin{equation} \label{eqSHAP3}
    \phi_{k,a}(x) = \sum_{\mathcal{S} \subseteq \mathcal{A} \setminus \{a\} } c(\mathcal{S},\mathcal{A}) \left[f_{k,\mathcal{S}\cup\{a\}}(x_{\mathcal{S}\cup\{a\}}) - f_{k,\mathcal{S}}(x_\mathcal{S})\right]
\end{equation}
being
\begin{equation*}
    c(\mathcal{S},\mathcal{A}) = \frac{|\mathcal{S}|!(|\mathcal{A}|-|\mathcal{S}|-1)!}{|\mathcal{A}|!}
\end{equation*}
and $\mathcal{A}$ is the set of all the features of the dataset, $a \in \mathcal{A}$ is the feature for which we want to compute the explanation value and $f_{k,\mathcal{S}}(x_\mathcal{S})$ is a function, built upon a partition $\mathcal{S}$ of the set of feature, that returns 1 if, for the data point $x_\mathcal{S}$ (which is the data point $x$ considering only the features in $\mathcal{S}$), the label $k$ is provided by the model (note that this particular notation makes explicit the fact that we are provided with a set of explanations, one for each feature, that are \emph{label specific}). The eq.~\ref{eqSHAP3} XAI function is linear with respect to the values predicted by the weak learners. Thereof, averaging the explanations produced by the weak learners provides the same value as directly applying the explanation algorithm to the entire ensemble which, in turn, produces explanations based on predictions that are the result of averaging the predictions of the decision trees that compose it (see supplementary material for further details).
Fig. \ref{fig:figure9} shows a toy example of how the RF regions of explanation constancy are constructed from decision trees weak learners resulting in smoother regions. Fig. \ref{fig:figure3} show a comparison on the variation of the value of the explanations for the different studied datasets. It can be observed that the heatmaps related to the explanations produced by Random Forest depict more gradual color changes, which correspond to differences in values, follow also a smoother (thus more desiderable) progression. For DT, some clear regions of explanations constancy are also observed.

\begin{figure}[!h]
\centering
\includegraphics[width=0.9\columnwidth]{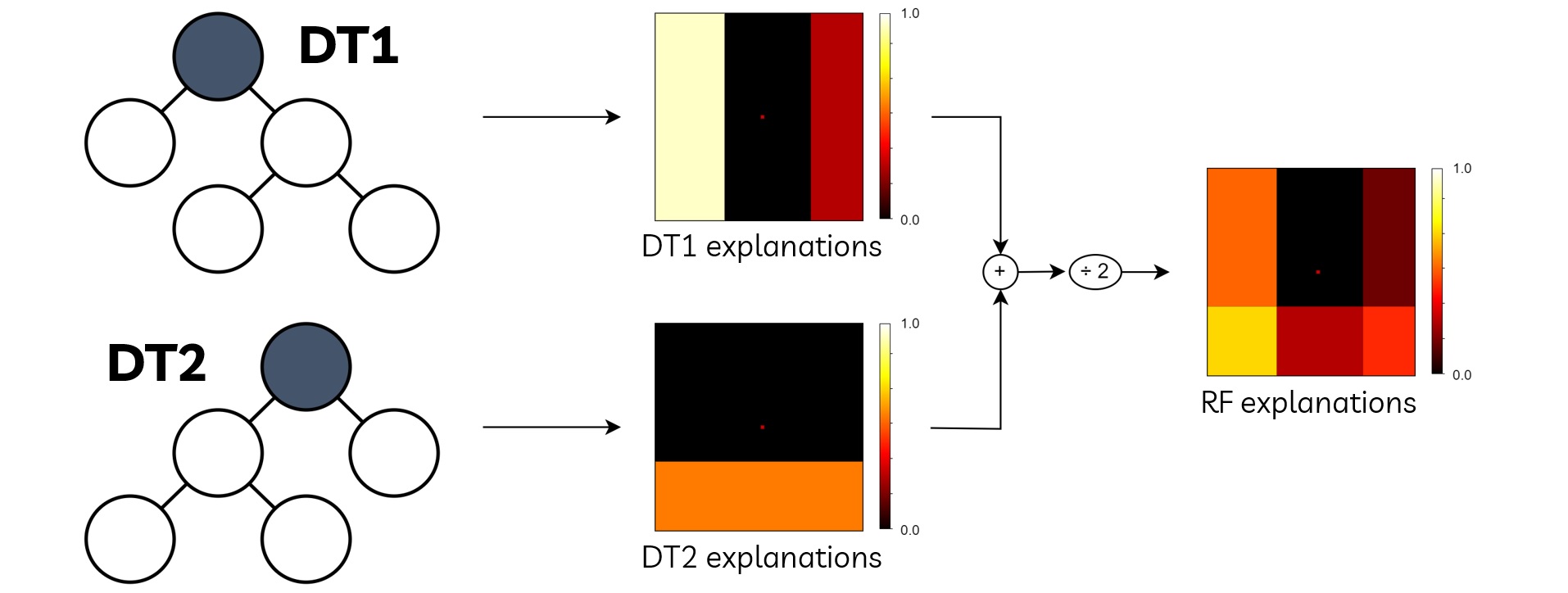}
\caption[From DT hetmaps to RF heatmaps]%
{\textbf{From DT heatmpas to RF heatmaps regions of explanation constancy} - Toy example of how an RF regions of explanation constancy are constructed from two DT models combined through averaging. The values represent the SHAP explanation difference function as defined in (\ref{heatmapsDifference}) (the point in red in the center is $x_i$, while the neighborhood is composed by the $x_j$ perturbed points). The RF values are significantly smoother, thus reflecting a more desirable behavior in terms of algorithm robustness.}
\label{fig:figure9}
\end{figure}

\begin{figure}[!h]
\centering
\includegraphics[width=\columnwidth]{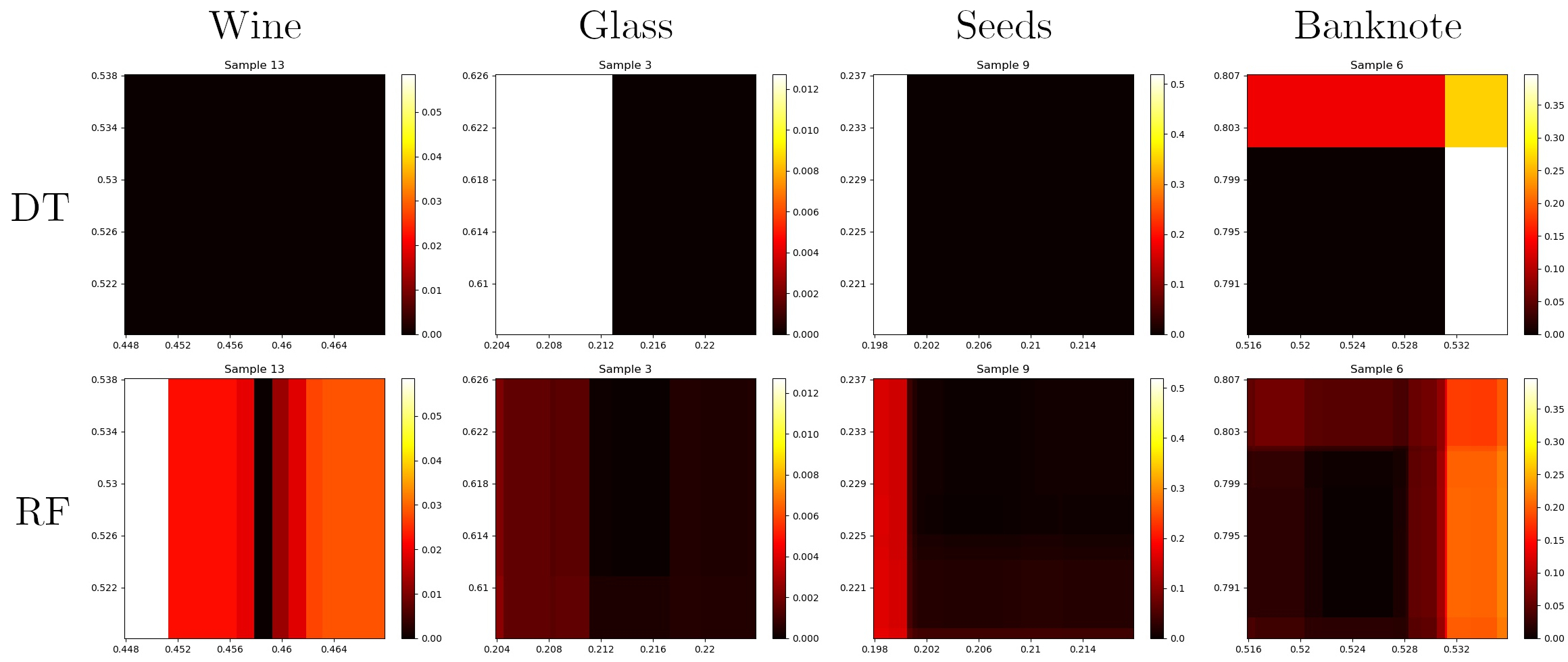}
\vspace{-5mm}
\caption[DT-RF comparison between heatmaps of SHAP explanations difference]%
{\textbf{DT-RF comparison between heatmaps of SHAP regions of explanation constancy} - Comparison between DT and RF regarding differences in the SHAP explanation values (see Eq. (\ref{heatmapsDifference})) around some points of interest of the different datasets.}
\label{fig:figure3}
\end{figure}

As we proof, application of SHAP to random forest improves the robustness over the use of a single decision tree but if we combine (by averaging) the decision tree week learners SHAP values then, we obtain the same results as if we apply SHAP technique to the random forest itself. Is there any other way that weak learners SHAP explanations (\emph{weak explanations}) can be combined to improve robustness? 
When SHAP is applied to the random forest model, it averages the SHAP values over the whose set of weak learners independently of they are right or wrong. This means that some of the \emph{weak explanations} are based on a model that may not be correct and thereof it is adding noisy explanations to the overall final explanation. To avoid this behavior, we "reward" those explanations from the models that contributed positively to the final decision. Within these terms, it can be considered to average only the weak explanations that come from the weak learners who provided as output the same label as the ensemble output. It is expected that a weak learner who contributed positively to the final ensemble output will be able to provide more relevant explanations regarding the decision made. The next section will be devoted to an in-depth analysis of this solution, called AXOM.

\subsection{Averaging on the eXplanations Of the Majority (AXOM)}
\begin{figure}[!tb]
\centering
\includegraphics[width=\columnwidth]{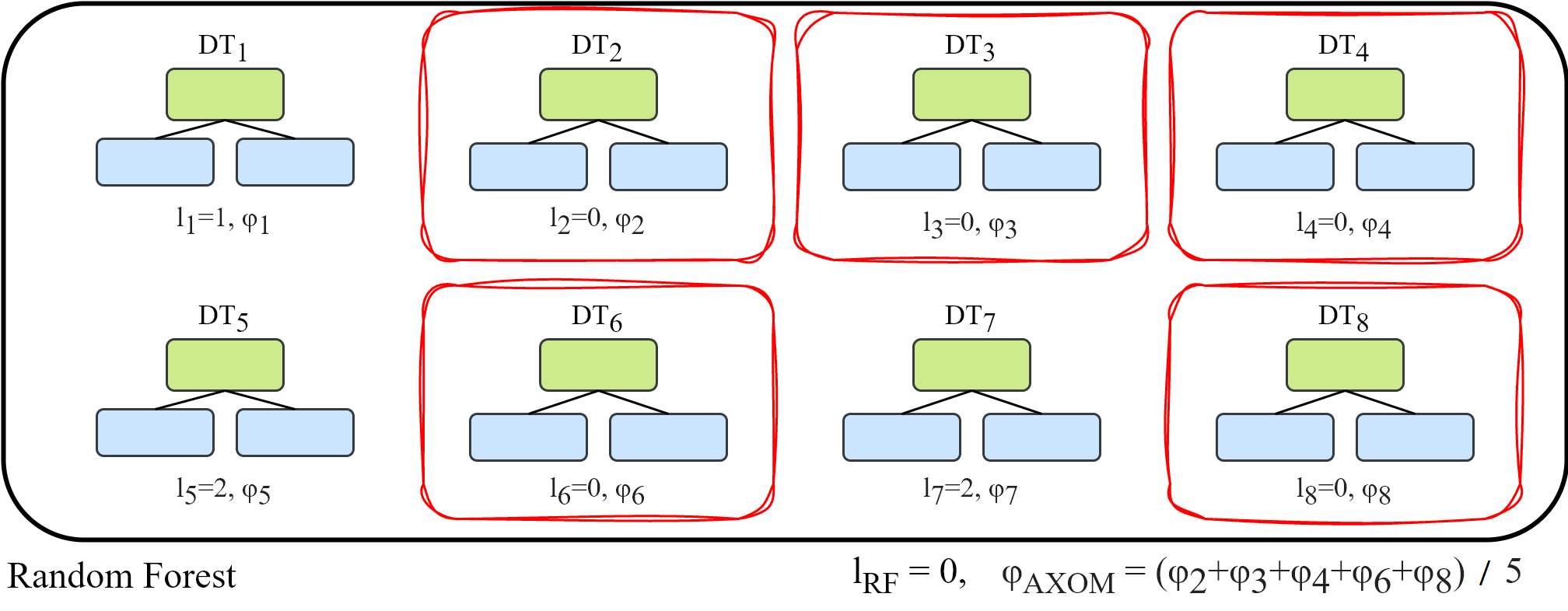}
\vspace{-5mm}
\caption[AXOM functioning illustration]%
{\textbf{AXOM functioning illustration} - A toy example to illustrate the functioning of AXOM: the ensemble (RF) consists of eight weak learners (DT) each of whom casts a vote $l_i$ for a given input, and has an associated explanation $\varphi_i$. Ensemble output is chosen according to a majority vote, in which $l_{RF}=0$ wins. To generate the output explanation, AXOM averages only the explanations of the weak learners who were part of the majority ($DT_2, DT_3. DT_4, DT_6, DT_8$), hence $\varphi_2, \varphi_3, \varphi_4, \varphi_6, \varphi_8$}.
\label{fig:figure8}
\end{figure}

The high explicability of the weak learners that make up the ensemble is certainly the most appealing property when it comes to trustworthiness of explanations. However, it must be considered that XAI algorithms such as SHAP, that are based on model output values, are highly dependent on the prediction accuracy for a given dataset. Low accuracy in the predictions of weak learners will produce explanations that are less robust. AXOM proposes to combine the explanations of individual learners of an ensemble to improve robustness and explainability. Specifically, we consider only to combine the explanations of weak learners that \emph{positively} contribute to the ensemble output (by \emph{positively} we mean that the classification output of the weak learner matches the one obtained by the random forest ensemble (see Fig. \ref{fig:figure8}).

In algorithm \ref{alg:alg1} the AXOM evaluation algorithm for each data point $x$ is presented.
The method receives as parameters the ensemble $e$ (a Random Forest trained model), the data point $x$ and the SHAP explainer $\sigma$. $\phi_w \in \mathbb{R}^{1 \times p}$ contains the SHAP explanations of the weak learner $w$, being $p$ the number of features, and an explanation is added to $\Phi \in \mathbb{R}^{n \times p}$, being $n$ the number of selected weak learners, if the label provided by the weak learner $l_w$ is equal to that predicted by the ensemble $l_e$. The final explanation $axom\_shap$ $\in \mathbb{R}^{1 \times p}$ is the mean of all selected weak explanations $\Phi$ for sample $x$.
This method ensures that the obtained explanation is associated with the final decision made by the ensemble. When a sample-specific explanation is produced, we expect to obtain data that support the decision that was actually made. Such information is most clearly extractable from the majority weak learners.

\begin{algorithm}[!h]
\caption{AXOM procedure to calculate single-sample explanations for an ensemble}
\label{alg:alg1}
\begin{algorithmic}
\Procedure{axom\_shap\_explanation}{$e,~x,~\sigma$}
    \State $l_e \gets e$.\Call{predict}{$x$}
    \Comment{Ensemble label prediction for $x$}
    \State $W \gets e$.estimators
    \Comment{Store the ensemble's weak learners set}
    \State $\Phi \gets$ \textbf{new} \Call{List}{ }
    \For{$w$ \textbf{in} $W$}
        \State $l_w \gets w$.\Call{predict}{$x$}
        \Comment{Weak learner label prediction}
        \If{$l_w = l_e$}
            \State $\phi \gets $\Call{shap\_explanation}{$w,x,\sigma$}
            \State $\Phi$.\Call{append}{$\phi$}
        \EndIf
    \EndFor
    \State $axom\_shap$ $\gets \frac{1}{|\Phi|}\sum_{\phi \in \Phi}\phi$
    \Comment{SHAP weak explanations mean}
    \State \Return $axom\_shap$
\EndProcedure
\end{algorithmic}
\end{algorithm}

\subsection{Datasets}
The methods have been tested on four commonly used classification datasets from \textbf{UCI Machine Learning Repository}: \textbf{\textsc{Wine}} \cite{WineDS} (3 classes), \textbf{\textsc{Glass Identification}} \cite{GlassDS} (7 classes), \textbf{\textsc{Seeds}} \cite{SeedsDS} (3 classes) and \textbf{\textsc{Banknote Authentication}} \cite{BanknoteDS} (2 classes). The table \ref{tab:datasets} shows some specific information as well as the accuracy of DT and RF models are specified.

%It is important to mention that among these four datasets, each one enjoys a good balancing of the classes, except for \textsc{Glass Identification}. Indeed, classes 0 and 1 alone represent almost the 70\% of the samples, while there are no samples at all belonging to class 4.

The number of features of the tested datasets was limited to 13. The reason behind this choice is related to computational power needs. In particular, given the choice of using 10000 points to evaluate the robustness around the neighbourhood area of interest, datasets with at most $\lfloor\log_2(10000)\rfloor = 13$ features were chosen in order to guarantee an adequate search in the entire feature space, that is, with at least two perturbations along each feature axis.

\begin{table}
\resizebox{\columnwidth}{!}{
\begin{tabular}{l|c|c|c|cc}
\multirow{2}{*}{\textbf{Datasets}} & \multirow{2}{*}{\textbf{N. of features}} & \multirow{2}{*}{\textbf{Training set size}} & \multirow{2}{*}{\textbf{Test set size}} & \multicolumn{2}{c}{\textbf{Accuracy}} \\
                                   &                                              &                                             &                                         & \textbf{DT}       & \textbf{RF}       \\ \hline
\textbf{\textsc{Wine}}                      & 13                                           & 160                                         & 18                                      & 88.9\%            & 100\%             \\
\textbf{\textsc{Glass}}                     & 10                                           & 192                                         & 22                                      & 81.8\%            & 95.5\%            \\
\textbf{\textsc{Seeds}}                & 7                                            & 189                                         & 21                                      & 85.7\%            & 95.2\%            \\
\textbf{\textsc{Banknote}}                  & 5                                            & 1234                                        & 138                                     & 98.6\%            & 99.3\%           
\end{tabular}
}
\vspace{-3mm}
\caption[Datasets description]%
{Descriptions of the datasets with accompanying information on DT and RF performance on them.}
\label{tab:datasets}
\vspace{-5mm}
\end{table}

\subsection{Experimental Design}
To conduct the test on robustness, the choice of the radius of the neighbourhood of the data points of the test set to be analysed was $\epsilon = 0.01$. This value defines the perturbation area to be analyzed and it is constant for all the experiments. It is important to mention that all data samples are normalised to 0-1 range and thereof the selected perturbation is of 1\%. The same experiments were performed for Decision Trees and Random Forest models. Both were trained on each of the four above-mentioned datasets to obtain the best model based on accuracy metric and by means of grid-search cross-validation. Two functions were then defined for calculating the value of $\bar{L}$, one that performs this calculation through the explanations obtained directly from the Decision Tree and Random Forest models and one that performs it on Random Forest through the previously defined AXOM algorithm. For each data point $x_i$ of the different test sets 10000 perturbed $x_j$ samples are randomly generated, on which the variation of the explanation value is calculated using Eq. (\ref{eqLipschitzAvg}). The method used for calculating the robustness of the explanations of a generic sample $x_i$ is reported In algorithm \ref{alg:alg2}.

\begin{algorithm}[!h]
\caption{Procedure to calculate mean robustness of explanation for a sample $x_i$}
\label{alg:alg2}
\begin{algorithmic}
\Procedure{compute\_mean\_robustness}{$e,x_i,~\sigma$}
    \State $\epsilon \gets 0.01$ 
    \State $n_{points} \gets 10000$
    \State $g_{x_i} \gets$ \Call{explain}{$e,x_i,\sigma$}
    \Comment{Explanation of $x_i$}
    \State $M \gets$ \textbf{new} \Call{List}{ }
    \State $\mathcal{N}_{f,\epsilon} \gets$ \Call{grid}{$x_i,\epsilon,n_{points},e$}
    \Comment{Neighborhood of 10k points}
    \For{$x_j$ \textbf{in} $\mathcal{N}_{f,\epsilon}$}
        \State $g_{x_j} \gets$ \Call{explain}{$e,x_j,\sigma$}
        \Comment{Explanation of $x_j$}
        \State $\mu \gets \frac{||g_{x_i} - g_{x_j}||_2}{||x_i - x_j||_2}$
        \Comment{Incremental ratio of $g_{x_i}$ and $g_{x_j}$}
        \State $M$.\Call{append}{$\mu$}
    \EndFor
    \State $\bar{L}$ $\gets \frac{1}{|M|}\sum_{\mu \in M}\mu$ \Comment{Overall robustness: average of all $x_j$}
    \State \Return $\bar{L}$
\EndProcedure
\end{algorithmic}
\end{algorithm}

The method receives as parameters the model $e$ (Decision Tree or Random Forest), the data point of interest $x_i$ and the SHAP explainer $\sigma$. First of all, the search parameters $epsilon$ and $n_{points}$ are fixed (in the algorithm we reported the values used in our experiments) and the SHAP explanation $g_{x_i}$ for point $x_i$ is calculated. After that, we build a grid $\mathcal{N}_{f,\epsilon}$ around the data point $x_i$ which is going to be used to carry out the evaluation of the robustness in the neighborhood (we recall that the neighborhood is composed by the $x_j$ point with same label as $x_i$, see Section \ref{sec:metric} for its definition). For each $x_j \in \mathcal{N}_{f,\epsilon}$ the SHAP explanation $g_{x_j}$ is produced and used to evaluate $\mu$, that is the variation of explanation value, as defined in Eq.(\ref{eqLipschitzAvg}). Finally, the mean robustness is calculated as the average of all the $\mu$ values. Note that this procedure is valid for both classic and AXOM SHAP explanations. Indeed, the method \textsc{explain} refers in a general way to the method for which one needs to asses the robustness.

\section{Results} \label{chap:results}
\begin{figure}[!tb]
\centering
\includegraphics[width=\columnwidth]{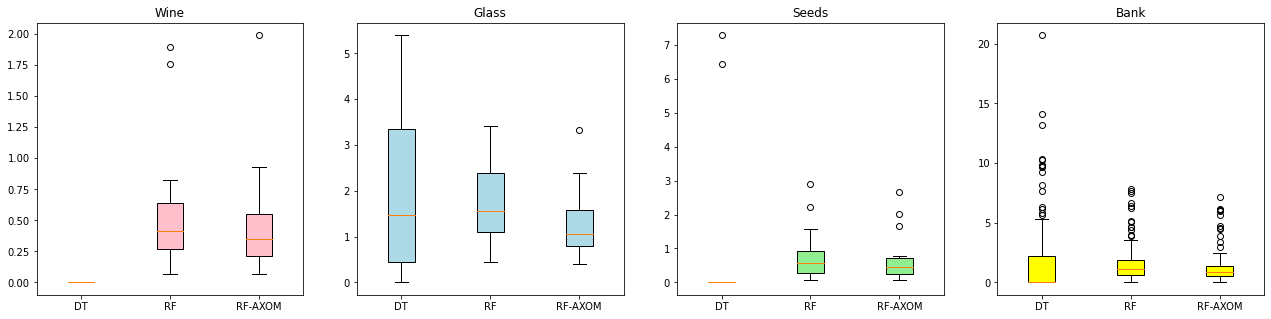}
\vspace{-5mm}
\caption[Robustness comparison through box plots]%
{\textbf{Robustness comparison through box plots} - Box plots constructed from the $\bar{L}(x_i)$ values of the $x_i$ samples of the test sets of each of the four analysed datasets. }
\label{fig:figure4}
\end{figure}

\textbf{Robustness comparison.} Table~\ref{tab:robustness} shows the $\bar{L}$ results in the form of mean and standard deviation for each model and dataset, while Fig.~\ref{fig:figure4} present them in a more detailed way by means of box plots (\textbf{Note}: $\bar{L}$ indicates the variation of explanation values in the neighborhood, thus a lower mean value is associated with a higher robustness).
It is possible to observe from the mean and standard deviation values that the AXOM procedure provides explanations that on average are more robust for all four analyzed datasets (for \textsc{Seeds} it has the same average robustness but the standard deviation is indicative of a better reliability of AXOM explanations). There is only one case in which this is breached at DT vs. AXOM and Wine dataset. DT overperforms AXOM, enjoying a seemingly perfect robustness despite its model accuracy is lower. This anomalous behavior is related with the complexity of the models and will be discussed later on this section.

\begin{table}
\resizebox{\columnwidth}{!}{
\begin{tabular}{l|cc|cc|cc|cc}
\multirow{2}{*}{\textbf{Model}} & \multicolumn{2}{c|}{\textbf{\textsc{Wine}}} & \multicolumn{2}{c|}{\textbf{\textsc{Glass}}} & \multicolumn{2}{c|}{\textbf{\textsc{Seeds}}} & \multicolumn{2}{c}{\textbf{\textsc{Banknote}}} \\
                                & Mean          & Std. Dev.          & Mean           & Std. Dev.          & Mean             & Std. Dev.             & Mean            & Std. Dev.           \\ \hline
\textbf{Decision Tree}          & 0.00          & 0.00               & 1.89           & 1.78               & 0.65             & 2.02                  & 1.75            & 3.32                \\
\textbf{Random Forest}          & 0.55          & 0.51               & 1.75           & 1.87               & 0.77             & 0.72                  & 1.58            & 1.57                \\
\textbf{AXOM}                   & 0.47          & 0.44               & 1.27           & 0.72               & 0.65             & 0.67                  & 1.28            & 1.34               
\end{tabular}
}
\vspace{-3mm}
\caption[$\bar{L}$ robustness comparison]%
{Mean and standard deviation of the $\bar{L}(x_i)$ values calculated for each sample $x_i$. Lower values of $\bar{L}$ denote better robustness to perturbations.}
\label{tab:robustness}
\end{table}

\textbf{Statistical significance.} To verify the reliability of these results, two-sample t-test for equal means was used (Student's T-tests was carried out for \textsc{Banknote} dataset that is normally distributed and Wilcoxon's T-tests was carried out for the others since the distributions were not normal). Table~\ref{tab:pvalues} shows the results. AXOM significantly improves robustness over RF for all datasets (see row RF vs. AXOM), with p-values all below the 0.05 threshold. The DT vs. AXOM results show a significant improvement for Seeds due to the improvement in the robustness standard deviation (see Fig.\ref{fig:figure4}). AXOM also performs better on average robustness value at Glass and Banknote datasets but the p-values are less significant being 0.113 and 0.066 respectively (we are neglecting for the moment the case of the \textsc{Wine} dataset which we will discuss later). Moreover, it can be observed that if we compare using AXOM or not for RF, the results for DT vs. AXOM are always better than the DT vs. RF. This indicates the effectiveness of the procedure.

\begin{table}
\centering
\resizebox{\columnwidth}{!}{
\begin{tabular}{l|cccc}
\multirow{2}{*}{\textbf{Comparison}} & \multicolumn{4}{c}{\textbf{p values}}                               \\
                                     & \textbf{\textsc{Wine}} & \textbf{\textsc{Glass}} & \textbf{\textsc{Seeds}} & \textbf{\textsc{Banknote}} \\ \hline
\textbf{DT vs. RF}                     & \textbf{<0.001}      & 0.774          & \textbf{0.008} &  0.3023             \\
\textbf{DT vs. AXOM}                   & \textbf{<0.001}      & 0.113          & \textbf{0.008} & 0.0656             \\
\textbf{RF vs. AXOM}                   & \textbf{0.042}       & \textbf{0.007} & \textbf{0.030}& \textbf{0.0444}         
\end{tabular}
}
\vspace{-3mm}
\caption[Two samples mean T-tests]%
{T-test values comparing RF, DT and AXOM robustness, for each dataset. In bold the p-values below 0.05.}
\label{tab:pvalues}
\end{table}

\textbf{Discussion.} 
Overall, more complex learners will increase accuracy (and vice-versa) and thereof the variance in the results~\cite{Black} which at last will also decreases robustness. Following the terminology used in that work it is stated that the prediction multiplicity (that is a subset of the procedure multiplicity) defined as models that have similar accuracy can predict different outputs, can be bounded by: $\frac{1}{2}V(M)\le I(M) \le 2V(M)$, being $V$ the variance of a set of possible models and $I(M)$ the predictive multiplicity. We introduce this terminology to relate that work with the presented approach and get insights into why it improves the explanation robustness criteria. In AXOM we limit the number of possible models to be considered by selecting those that have the same output than the global RF model. By doing this, we are reducing the variability $V(M)$ of the possible chosen models $M$ (under the random models distribution created by random selection of features and samples) and thereof mitigating the predictive multiplicity problem. This is only true if the global output is correct (otherwise we are considering the 'wrong' weak learners) and the reason why robustness of AXOM is also linked to the accuracy of the Random Forest model (the higher the accuracy of the global model is the higher the expected robustness of AXOM method).

%we present two graphical illustrations by means of heatmaps constructed according to the incremental ratio of explanation values within the neighborhood of the $x_i$ points of interest.

\begin{figure}[!tb]
\centering
\includegraphics[width=\columnwidth]{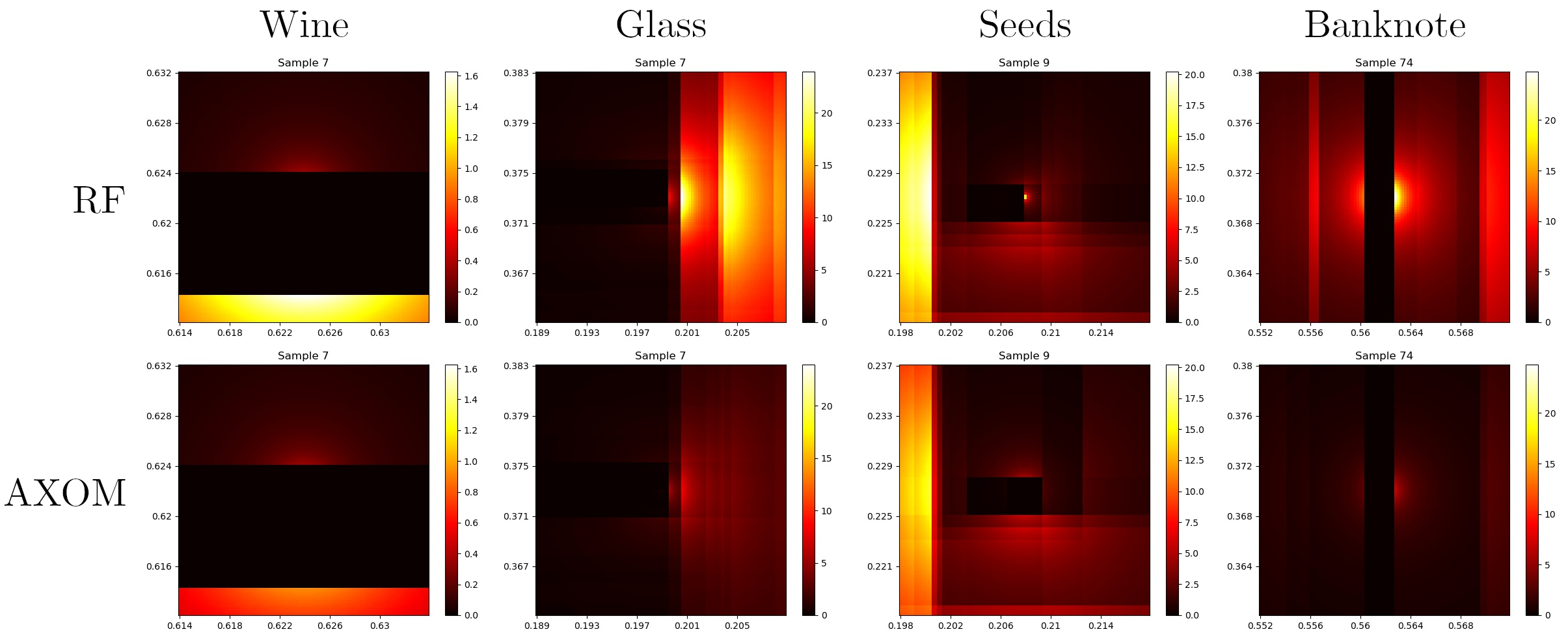}
\vspace{-5mm}
\caption[RF and AXOM Robustness incremental ratio heatmaps]%
{\textbf{RF and AXOM Robustness incremental ratio heatmaps} - Sample-specific comparison between the robustness heatmaps of explanations (see eq (\ref{heatmapsRobustness})) of RF (top) and AXOM (bottom). RF and AXOM produce explanations that vary proportionally equally (although AXOM values are smaller) except for the absence of some boundaries in AXOM.}
\label{fig:figure2}
\end{figure}

\begin{figure}[!t]
\centering
\includegraphics[width=\columnwidth]{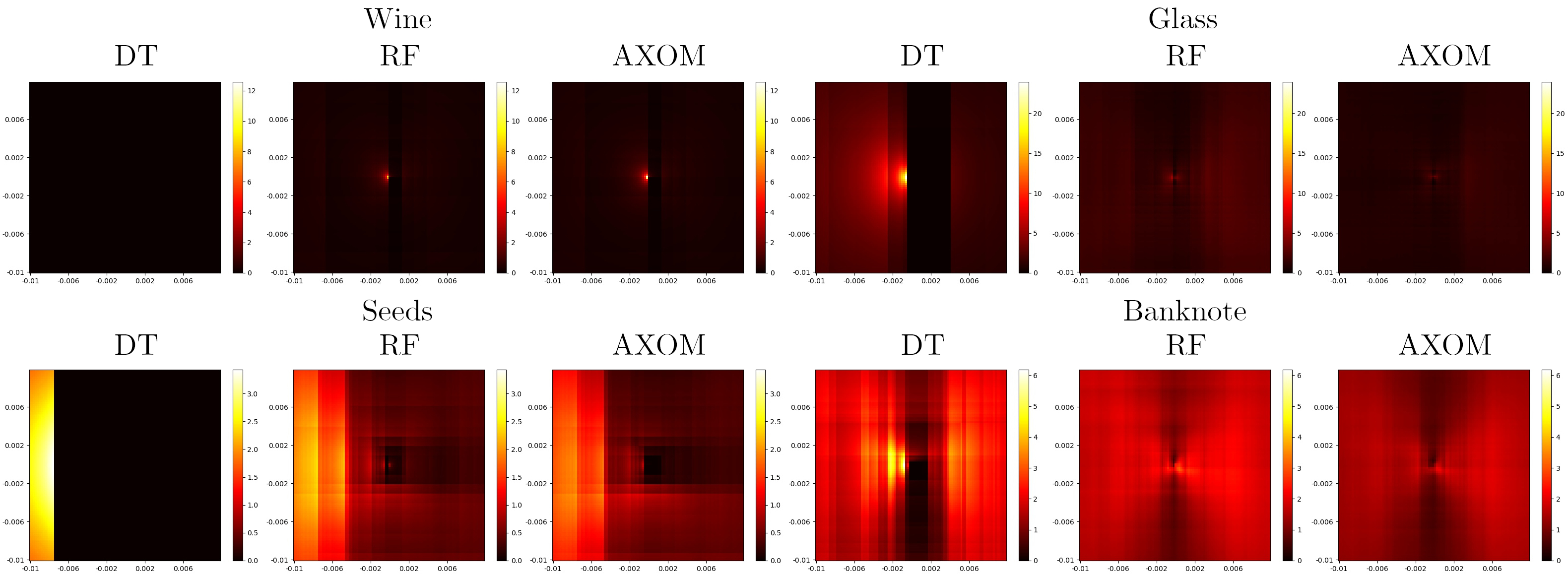}
\vspace{-5mm}
\caption[DT, RF, and AXOM full test set Robustness comparison]%
{\textbf{DT, RF, and AXOM full test set Robustness comparison} -Robustness heatmaps of the explanations (see eq (\ref{heatmapsRobustnessFullD})) for DT, RF and AXOM and the full dataset. The plot was produced by centering in (0, 0) all the $x_i$ samples in the dataset so that all the samples could be fit into the same box of size $2\epsilon \times 2\epsilon$. RF and AXOM enjoy better smoothness and robustness in value changes than DT, with AXOM in particular possessing darker plots in color than RF, indicative of better average value robustness.}
\label{fig:figure10}
\end{figure}

To illustrate the results Fig. \ref{fig:figure2} shows the heatmaps of the robustness of the neighborhood of all four test sets, focusing on RF (top) and AXOM (bottom), taking a representative sample from each of the four datasets as an example. Each $H(x_m)$ value of the map is computed as defined in Eq. (\ref{heatmapsRobustness}). It can be observed that the RF and AXOM SHAP explanations vary similarly around the $x_i$ points and the main differences are represented by darker colors at AXOM, indicating smaller robustness variations. Fig. \ref{fig:figure10} shows a similar comparative (including also DT), but overlapping (through averaging) the heatmaps of all test samples. Being $\mathcal{T}$ the set of all test points of a given dataset the \emph{heat} value in correspondence of each generic $x_m$ point of the map is given by the following formula (based on Eq.~\ref{heatmapsRobustness}) :

\begin{equation} \label{heatmapsRobustnessFullD}
    H(x_m) = \frac{1}{|\mathcal{T}|}\sum_{x_i \in \mathcal{T}} \frac{||g(x_i) - g(x_j(x_m))||_2}{||x_i - x_j(x_m)||_2}
\end{equation}

%To be more specific, being $\mathcal{T}$ the set of all test points of a given dataset, all $x_i \in \mathcal{T}$ test samples were centered in (0, 0) so that all samples could be fit in the same box with axes bounded inside $(-\epsilon, \epsilon)$, in which the $x_m$ points vary. As explained in Section \ref{sec:metric}, there is a correspondence between $x_j$ and $x_m$, this time slightly differently defined as:
%\begin{align*}
%    x_j(x_m) = \{ & x_{j,1},...,x_{j,a_x-1},x_{j,a_x}+x_{m,a_x},x_{j,a_x+1},..., \\
%    & x_{j,a_y-1},x_{j,a_y}+x_{m,a_y},x_{j,a_y+1},...,x_{j,p} \}
%\end{align*}

It is possible to see from the colors of the plots that AXOM (except for the usual anomalous case represented by the \textsc{Wine} dataset) always exhibit more desirable behavior than DT and RF, with significantly smaller explanation values that also vary more smoothly.

We previously considered that there is an anomalous behavior for DT and Wine dataset regarding its robustness value. Indeed, it can be observed from Table \ref{tab:robustness} that the obtained robustness is 0. That means that the robustness is "perfect", i.e. all the SHAP explanations for the perturbed data have exactly the same value as the original data point. 
%However, the production of explanations that are constant over a large portion of the feature space is only desirable behaviour if the problem to be explained is simple, which represents a contradiction, as the need for explanations grows with the complexity of the problem. This explains the need to construct explanations in such a way that they are capable of modelling more complex behaviour and thus, analogous to matters concerning the accuracy of a model, justifies the possibility of constructing "ensembles of explanations".
To recall the constancy of the explanations, at Fig.~\ref{fig:figure3} can be observed that for \textsc{Wine} dataset the differences in explanations of DT are zero-valued (explanation values are constant) which means that all perturbed points  fall into the same branch of the sample. Indeed, it can be deduced from the box plots (see Fig.~\ref{fig:figure4}) that DT tends to provide explanations that enjoy perfect robustness only in samples that are sufficiently distant (i.e. far enough away not to be affected by the perturbations) from the decision surfaces where a branch change occurs. Thereof, DT presents this behavior when a simple and clearly separated decision boundaries are learned. This is not a bad by itself as far as accuracy is preserved, but at table~\ref{tab:datasets} can be seen that the accuracy is lower (88.9\% vs. 100\%). Note that there is always a balance between model complexity and robustness of explanations independently of the specific model or XAI technique used. 

%An example of this behaviour is shown in Fig. \ref{fig:figure7} through two representative samples of \textsc{Wine} test set. Specifically, by setting $\epsilon=0.2$ (in the experiments $\epsilon=0.01$ was set), and thus enlarging the range of points plotted in the heatmaps concerning the difference in explanations as defined in (\ref{heatmapsDifference}), it can be seen that DT provides explanations that, although enjoying (apparent) perfect robustness in a relatively large neighbourhood, suffer abrupt changes in value as soon as a change in the decision branch is reached, with values significantly higher than those of AXOM. We recall that in DT a change of decision branch does not necessarily imply a change of predicted label. In fact, according to the algorithm used to produce the heatmaps, if such areas appeared in this plot they would be characterized by the color grey.

%\begin{figure}[!b]
%\centering
%\includegraphics[width=\columnwidth]{figure7.jpg}
%\vspace{-5mm}
%\caption[DT-AXOM explanations smoothness comparison in Wine dataset]%
%{\textbf{DT-AXOM explanations smoothness comparison in Wine dataset} - DT and AXOM explanations difference heatmaps for two representative samples of \textsc{Wine} dataset. The plots were produced using a maximum perturbation $\epsilon=0.2$ to show that tweaking a data point enough to change decision branch of DT, results in abrupt value changes in the SHAP explanations. This proof that DT only enjoys \emph{apparent} good robustness.}
%\label{fig:figure7}
%\end{figure}

\begin{table}
\centering
\begin{tabular}{lcccc}
                                                                                                    & \textbf{Wine} & \textbf{Glass} & \textbf{Seeds} & \textbf{Banknote} \\ \cline{2-5} 
\multicolumn{1}{c|}{\textbf{\begin{tabular}[c]{@{}c@{}}Weak Mislabeling\\ Percentage\end{tabular}}} & 12.1\%        & 24.9\%         & 14.0\%         & 2.5\%            
\end{tabular}
\vspace{-3mm}
\caption[Weak learner's mislabeling percentage]%
{The values represent the average number of weak learners that cast a vote different with respect to the final ensemble prediction.}
\label{tab:table3}
\vspace{-3mm}
\end{table}

\begin{figure}[!tb]
\centering
\includegraphics[width=\columnwidth]{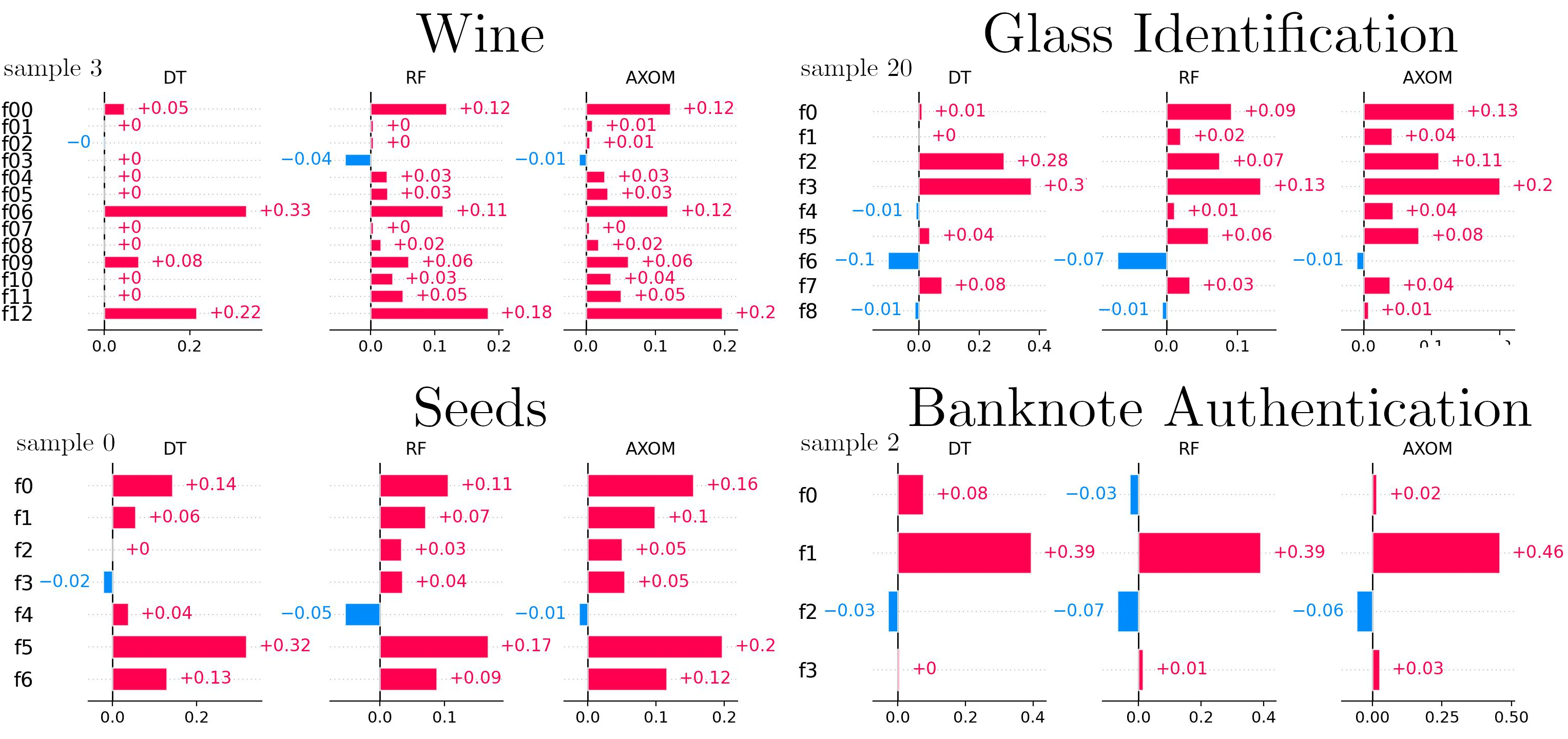}
\vspace{-5mm}
\caption[Sample-based SHAP values comparison]%
{\textbf{Sample-based SHAP values comparison} - Comparison of the SHAP values produced by DT, RF and AXOM to explain a representative sample of each of the datasets. Differences can be observed at some values such as Banknote F0 or Glass F4 features.}
\label{fig:figure5}
\end{figure}

\begin{figure}[!tb]
\centering
\includegraphics[width=\columnwidth]{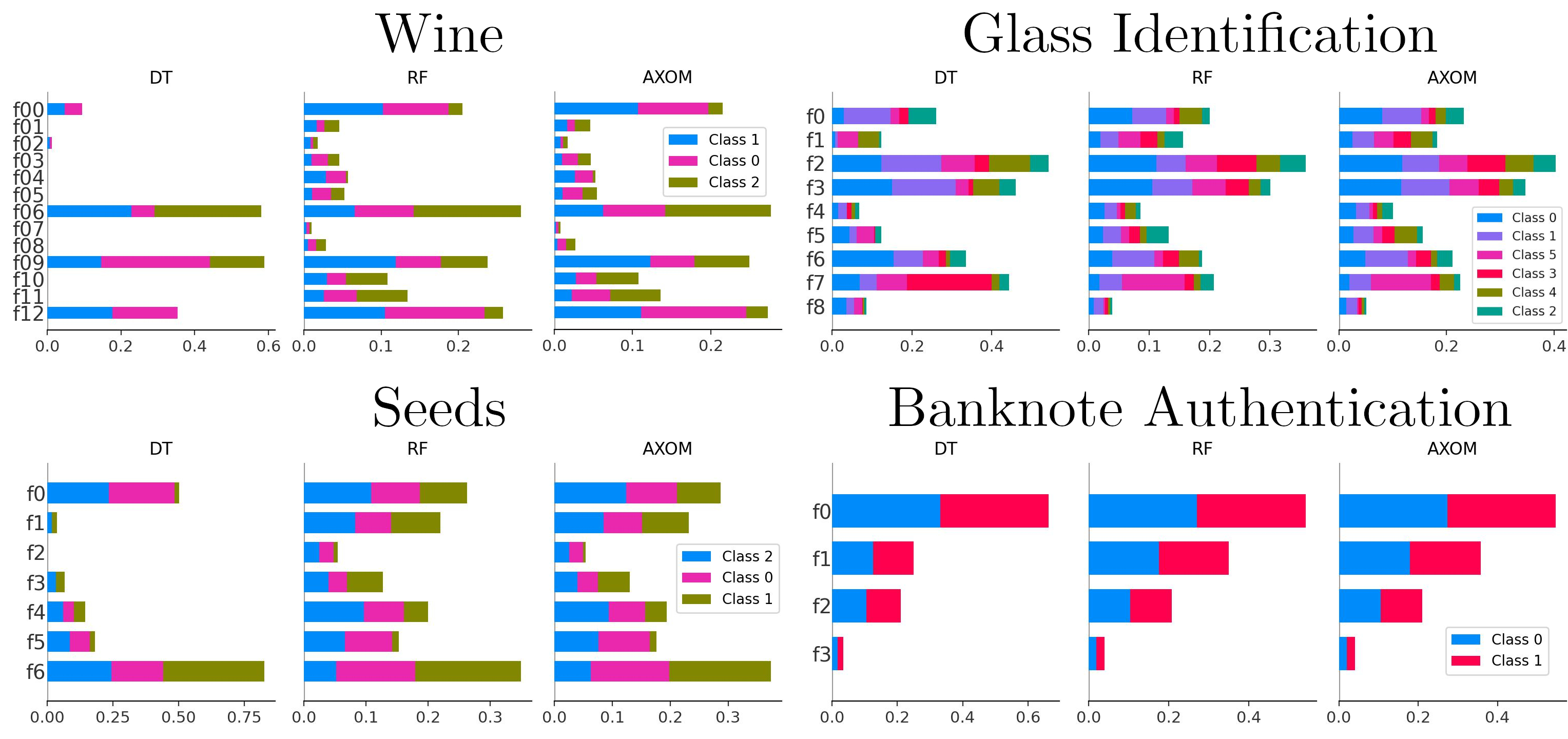}
\vspace{-5mm}
\caption[Full test set SHAP values comparison]%
{\textbf{Full test set SHAP values comparison} - Comparison of the multi-output SHAP explanations produced by DT, RF and AXOM.}
\label{fig:figure6}
\end{figure}

\textbf{SHAP values comparison.} AXOM method improves robustness by selecting a subset of weak learners obtaining a new set of SHAP values and hence, a modified set of explanations. Fig. \ref{fig:figure5} shows the explanations produced from the predicted labels of four representative samples of the different test sets, while Fig. \ref{fig:figure6} shows the multi-label explanation values of the entire test sets. As can be observed, RF and AXOM tend to distribute the "merit" of the produced output across all features, modeling and explaining more complex behaviours, compared to DT which tends to load as much responsibility as possible into fewer features. Note that AXOM and RF explanations will be more similar if all the weak learners outputs match with the ensemble output. In this regard, Table \ref{tab:table3} shows the average percentage of weak learners who provided a label different from the ensemble on the test data, i.e., the percentage of weak explanations discarded by AXOM. It is possible to see that \textsc{Glass} is the dataset in which there is more indecision within voting, while in \textsc{Bank} we observe behavior tending toward voting unanimity, which is reflected respectively in a lower and higher similarity of explanations, as observable in Fig. \ref{fig:figure6}. For example, it can be seen that in \textsc{Glass} the explanations of classes 2 and 4 differ significantly between RF and AXOM for almost all features, in addition to the fact that the ranges of values are higher in the case of AXOM.

 %which was initially used as a starting point to formally assess the robustness of the classic model explanations, and subsequently as a key tool to conduct a comparative analysis between the latter and the developed procedures. The comparisons immediately showed that combining the explanations of the weak learners in an ensemble leads to the production of values that significantly improve this characteristic, so it was decided to explore such a path until the development of the procedure called AXOM (Averaging on the eXplanations Of the Majority), with which remarkable results were obtained.

\section{Conclusions}
There is a growing concern about the robustness of some interpretability frameworks and how to guarantee that the robustness of the explanations is, at least, as robust as the underlying model. In this work we set to investigate whether weak learners explanations can be combined to improve robustness of explanations for ensemble models. We introduced a modified version of a robustness of model explanations criteria and proposed an algorithm for calculating the SHAP explanations of ensembles models as the result of averaging the explanation SHAP-values of weak learners who contributed positively to the final prediction. This approach has proven to be a method that significantly improves the robustness of model explanations compared to explanations obtained through the direct application of XAI methods to the ensemble under consideration. Removing the explanations of weak learners whose prediction is not consistent with the global ensemble, it is also reduced the model variance alleviating the predictive multiplicity problem without penalizing the accuracy. We can confirm that the use of weak explanations can play a key role in explaining the decisions of the ensemble and the presented approach may not be limited to Random Forest and SHAP, but it could be extended to other ensemble models, such as Bagging or eXtreme Gradient Boosting, and other post-hoc XAI techniques.

\ack Tthis work has been supported by the IAX (eXplainable Artificial Intelligence) grant funded by the Comunidad de Madrid Jóvenes Investigadores 2022/2024 initiative.

\bibliography{ecai}
\end{document}